\pdfoutput=1

\documentclass[11pt]{article}

\usepackage[final]{acl}

\usepackage{times}
\usepackage{latexsym}
\usepackage{array}

\usepackage[T1]{fontenc}

\usepackage[utf8]{inputenc}
\usepackage{CJK}
\usepackage{booktabs}
\usepackage{microtype}
\usepackage{arydshln}
\usepackage{inconsolata}

\usepackage{graphicx}
\usepackage{xcolor}
\usepackage{makecell}
\usepackage{enumitem}
\usepackage{amsmath}
\usepackage{multirow}
\usepackage{xspace}
\usepackage{dashrule}

\definecolor{forestgreen}{rgb}{0.13, 0.55, 0.13}

\title{Marco-o1 v2: Towards Widening The Distillation Bottleneck for Reasoning Models}

\author{
    Huifeng Yin\textsuperscript{1, 2\thanks{Equal contribution.}}~
    Yu Zhao\textsuperscript{1\footnotemark[1]}~
    Minghao Wu\textsuperscript{1, 3}~
    Xuanfan Ni\textsuperscript{1}~
    Bo Zeng\textsuperscript{1}~
    Hao Wang\textsuperscript{1}~
    Tianqi Shi\textsuperscript{1}~\\
    \textbf{Liangying Shao\textsuperscript{1}~
    Chenyang Lyu\textsuperscript{1}~
    Longyue Wang\textsuperscript{1\thanks{Corresponding author.}}~
    Weihua Luo\textsuperscript{1\footnotemark[2]}~
    Kaifu Zhang\textsuperscript{1}~}\\
    \textsuperscript{1}Alibaba International Digital Commerce \\
    \textsuperscript{2}Tsinghua University ~~~~~~~~ \textsuperscript{3}Monash University\\
    {\tt \{fengli.zy, wanglongyue.wly, weihua.luowh, kaifu.zkf\}@alibaba-inc.com}
}

\begin{document}
\maketitle
\begin{abstract}
Large Reasoning Models~(LRMs) such as OpenAI o1 and DeepSeek-R1 have shown remarkable reasoning capabilities by scaling test-time compute and generating long Chain-of-Thought~(CoT). Distillation post-training on LRMs-generated data is a straightforward yet effective method to enhance the reasoning abilities of smaller models, but faces a critical bottleneck: we found that distilled long CoT data poses {\em learning difficulty} for small models and leads to the {\em inheritance of biases} (i.e. over-thinking) when using Supervised Fine-tuning~(SFT) and Reinforcement Learning~(RL) methods. To alleviate this bottleneck, we propose constructing tree-based CoT data from scratch via Monte Carlo Tree Search~(MCTS). We then exploit a set of CoT-aware approaches, including {\em Thoughts Length Balance, Fine-grained DPO}, and {\em Joint Post-training Objective}, to enhance SFT and RL on the constructed data.
We conduct evaluation on various benchmarks such as math~(GSM8K, MATH, AIME). instruction-following~(Multi-IF) and planning~(Blocksworld), results demonstrate our approaches substantially improve the reasoning performance of distilled models compared to standard distilled models via reducing the hallucinations in long-time thinking. The project homepage is \url{https://github.com/AIDC-AI/Marco-o1}. 

\end{abstract}

\section{Introduction}

Recent advancements in large reasoning models (LRMs), such as OpenAI o1~\cite{openai2024reason}, QwQ \cite{qwq-32b-preview} and DeepSeek-R1~\cite{guo2025deepseek}, have led to significant progress in handling complex tasks spanning mathematics, coding, and even open-ended queries~\cite{zhong2024evaluation,huang2024o1,zhao2024marco}. The success is largely attributed to ``scaling test-time compute'' by extending the length of the reasoning process. 
{Given that most state-of-the-art LRMs are computationally expensive, recent efforts attempt to distill their reasoning capabilities into smaller lightweight models, demonstrating competitive performances~\cite{qin2024o1}.} 
For instance, \citet{guo2025deepseek} explored {\em direct distillation}, where they fine-tuned smaller dense models (e.g. Qwen2.5 7B) using reasoning patterns generated by DeepSeek-R1 671B model, outperforming GPT-4 on math benchmarks (e.g. AIME: 9.3\% vs. 55.5\%).

\begin{CJK}{UTF8}{gbsn}
\begin{table*}[t]
\centering
\scalebox{0.8}{
\begin{tabular}{p{3.1cm} p{16cm}}
\toprule
\bf Error Type  & \bf Thoughts \\
\midrule
{\bf Content Repetition} \newline (Math)  & Five men and nine women stand equally spaced around a circle in random order. The probability that every man stands diametrically opposite a woman is $\frac{m}{n}$, where $m$ and $n$ are relatively prime positive integers. Find $m+n$.\\
\cdashline{2-2}
& So I've got this probability problem here. There are five men and nine women standing equally spaced around a circle in random order. I need to find the probability that $\dots$ 

$\dots$ 

The total number of circular permutations of 14 distinct people is (14 - 1)! = 13!, because in circular arrangements, one position is fixed to handle the rotational symmetry.So, total number of possible arrangements is 13!.

$\dots$ 

Wait, but in circular arrangements, positions are considered up to rotation, so fixing one position is standard. Alternatively, perhaps I need to consider that the {\color{blue} \em positions} are {\color{red} considered} up to {\color{red} consider} that the {\color{blue} \em positions} are {\color{red} considered} up to that the {\color{blue} \em positions} are {\color{red} considered} up to {\color{red} consider} that the {\color{blue} \em positions} are {\color{red} considered} up to $\dots$ {\color{forestgreen}\small (no answer)}\\
\midrule
{\bf Over-Reflection} \newline (Instruction-  & How to tell others that your major is computer science, without saying the word computer or science? You entire response should not contain the word computer and science.\\
\cdashline{2-2}
Following) & So I have this problem: I need to explain what my major is without saying "computer" or "science". Hmm, my major is in computer science, but I have to find another way to describe it.

$\dots$

Maybe I can describe it in terms of what I do. I learn how to design and develop software, understand how information is processed and stored, and how to solve problems using logical methods. I also study the theory behind it, like algorithms and data structures.

$\dots$

{\color{red} I should also consider} $\dots$ {\color{red} Alternatively, } $\dots$ {\color{red} I think another angle is} $\dots$ {\color{red} Wait, perhaps} $\dots$ {\color{red} I could say that} $\dots$ {\color{red} Or, I might} $\dots$ {\color{red} I think another way} $\dots$ {\color{red} Alternatively,} $\dots$ {\color{red} I should also consider} $\dots$ {\color{forestgreen}\small (no answer)}\\
\midrule
{\bf Instruction Failure} & Translate the following sentence into English: 您好，很高兴为您效劳，请问有什么可以帮您。\\
\cdashline{2-2}
(Translation) & Translation: {\color{blue} \em Okay}, I'm happy to help you. Do you have any questions for me?

{\color{red}Wait, that's not correct.} The original sentence is in Chinese, and the translation should be accurate. Let me think again.

The original sentence is: {\color{blue}您}, 很高兴为您效劳, 请问有什么可以帮您.

Breaking it down:

"您好" is "Hello" or "Hi". "很高兴为您效劳" translates to "I am very pleased to serve you." $\dots$

Wait, the original sentence is more like a greeting and offer to help. $\dots$

Alternatively, "Hello, I'm happy to help you. How can I assist you?" So perhaps the correct translation is: "Hello, I'm here to help you. How can I assist you?"

Wait, $\dots$

Alternatively, $\dots$
{\color{forestgreen}\small (no answer)}\\
\bottomrule
\end{tabular}}
\caption{The illustration of {\bf formalistic long-time thinking} generated by distilled reasoning models across different tasks. Error tokens in thoughts are highlighted in {\color{blue}\em blue} and {\color{red}red} colors. Notably, due to excessively long thoughts, there are {\color{forestgreen}\em no final answers} in above cases. The Quantitative Analysis is detailed in Section \ref{sec:sft_dpo_data}.}
\label{tab:formalistic}
\end{table*} 
\end{CJK}



However, we observed that these distilled models often exhibit hallucinations during long-time thinking, such as content repetition and over-reflection, leading to no final answer being produced (as shown in Table~\ref{tab:formalistic}). We refer to this phenomenon as {\bf formalistic long-time thinking}, where smaller models mechanically replicate the reasoning patterns of large models without internalizing the reasoning logic. 
Recent research shows that LRMs face both over-thinking and under-thinking issues \cite{chen2024not, wang2025thoughts}, while smaller models struggle to learn general reasoning \cite{fu2023specializing}.
Accordingly, the root cause may be that distillation methods introduce {\em bias inheritance} and {\em learning difficulties} in smaller models.
A natural research question arise: {\em How can long CoT reasoning be effectively transferred to smaller models through data construction, SFT and RL methods?} 



To tackle this challenge, we explore improvements in reasoning distillation from both data and methodological perspectives.
First, we propose a fundamental framework for constructing tree-based CoT data, which generates pre-defined thought nodes using general LLMs (rather than LRMs) and heuristically expands these nodes into a tree structure via the Monte Carlo Tree Search (MCTS) algorithm \cite{browne2012survey}. The constructed data is not only more effective compared to directly distilled data, but also inherently more flexible, allowing the extraction of different types of reasoning paths as training data.
Secondly, regarding commonly-used SFT and direct preference optimization (DPO) \cite{NEURIPS2023_a85b405e} as post-training framework, we empirically investigate a set of CoT-aware methods on the effects of formalistic long-time thinking. Specifically, this includes: 1) {\em Thoughts Length Balance}, where we extract CoT data of varying lengths; 2) {\em Fine-grained DPO}, where we employ conservative DPO (cDPO) \cite{mitchell2023note} and mask-based DPO to better leverage the fine-grained information in long CoT; 3) {\em Joint Post-training Objective}, where we combine the DPO loss with SFT loss to mitigate the over-optimization observed in DPO \cite{DBLP:journals/corr/abs-2410-15483,DBLP:journals/corr/abs-2410-21438}. 





We validated our approaches on five exam-oriented and open-ended benchmarks, covering three different difficulty levels of math (GSM8K, MATH and AIME) \cite{cobbe2021gsm8k,lightman2023lets}, instruction-following in eight languages (Multi-IF) \cite{he2024multi}, and real-world planning tasks (Blocksworld) \cite{Valmeekam2022PlanBenchAE}. Experimental results show that the proposed method consistently and orthogonally improve reasoning performance over the standard distilled models. The improvements come from the reduced hallucinations during long-time thinking, particularly content repetition, which leads to fewer ``no answer'' phenomena and better overall accuracy. 
The {\bf main contributions} of this work are:

\begin{itemize}[leftmargin=*,topsep=0.1em,itemsep=0.1em,parsep=0.1em]

\item Our study reveals the side effect of 
standard distillation on transferring long CoT reasoning, which results in sub-optimal training of smaller models when using the distilled data (in Section \ref{sec:sft_dpo_data}). 

\item {We propose a novel approach to construct CoT trees from scratch, which not only scales up the solution space but also more closely mimics human-like reasoning patterns. To the best of our knowledge, it is the first attempt of its kind (in Section \ref{sec:2}).} 


\item We investigate a set of effective approaches to widen the distillation bottleneck, demonstrating that they are orthogonal and complementary to each other, and robustly applicable to different reasoning tasks and languages (in Section \ref{sec:3}\&\ref{sec:4}).


\end{itemize}

\begin{figure*}[t]
\centering
\includegraphics[width=0.8\textwidth]{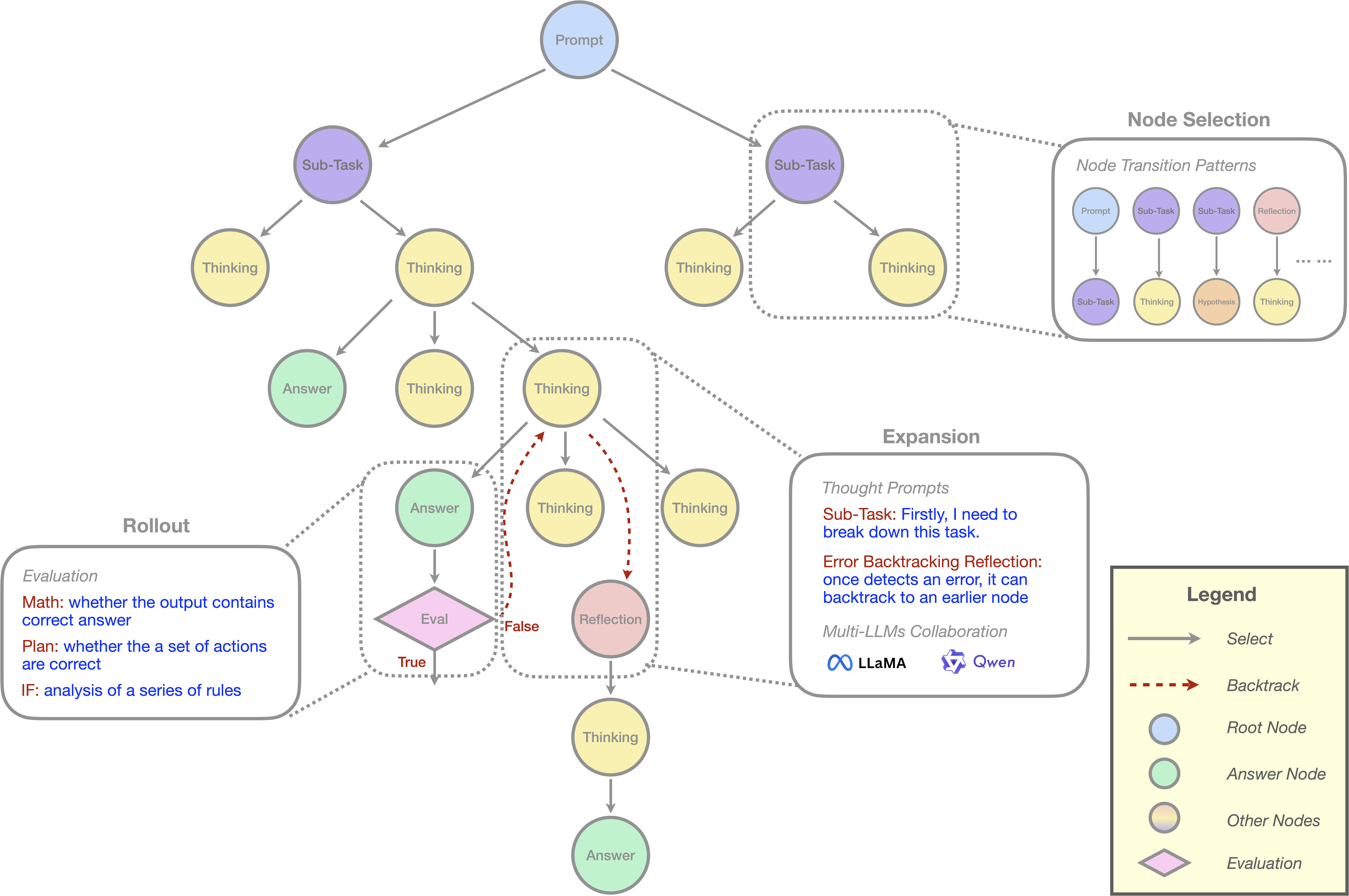}
\caption{MCTS-based CoT data generation framework. Starting from an initial prompt (root node), the system proceeds through predefined nodes (\emph{e.g.}, \textit{Sub-Task}, \textit{Thinking}, \textit{Reflection}) according to a customizable node transfer matrix. Each node is expanded by prompting either Qwen or Llama, allowing multi-model collaboration. If a wrong answer is detected, we perform error backtracking (pink arrows) to a prior node and trigger \textit{Reflection} in another model, enhancing the overall correctness and diversity of the final reasoning path.}
\label{fig:data_construction}
\end{figure*}

\section{Tree-Based CoT Data Construction}
\label{sec:2}

We propose a flexible and customizable tree-based CoT data construction method that generates high-quality CoT data from scratch. In this section, we introduce the overall framework in Section~\ref{sec:2.1}, followed by the thought nodes (Section~\ref{sec:2.2}), reasoning patterns (Section~\ref{sec:2.3}), and how to extract CoT data for post-training (Section~\ref{sec:sft_dpo_data}).

\subsection{Overall Framework}
\label{sec:2.1}
We introduce the tree-based CoT data construction process as shown in Figure~\ref{fig:data_construction}. 
This tree structure not only constrains the search space to prevent unbounded expansions but also guides the model to produce reasoning steps (nodes) systematically. For instance, we specify that each node in the search tree corresponds to a particular action role (\emph{e.g.}, \textit{thinking}, \textit{reflection}), and each edge represents a transition to the next step. By constraining the transitions among these nodes, we ensure the search is both tractable and coherent.
With the structure in place, we use MCTS to explore the search tree. During each step:
\begin{itemize}[leftmargin=*,topsep=0.1em,itemsep=0.1em,parsep=0.1em]
\item {\bf Node Selection.} We select a thought node to expand based on MCTS principles, such as upper confidence bound (UCB). If $\mathrm{Child}(n)$ denotes the set of child nodes of node $n$, then UCB balances exploration and exploitation via a score:
$UCB(n_i) = \frac{v(n_i)}{n_{\text{visits}}(n_i)} + C \sqrt{\frac{\ln{\bigl(n_{\text{visits}}(n_{parent})\bigr)}}{n_{\text{visits}}(n_i)}}$, 
where $v(n_i)$ is an estimated value (or reward) of node $n_i$, $n_{\text{visits}}(\cdot)$ denotes the visit count, and $C$ is the exploration constant.
    
\item {\bf Expansion.} We expand the selected node by prompting an LLM by adding a thought prompt (detailed in Table~\ref{tab:prefill_prompt}) that specifies the required action role. The LLM then generates the textual content for that node.

\item {\bf Rollout.} If the expansion reaches an \textit{answer} node, we compute a reward based on correctness determined by rules and backpropagate this reward up the tree.
\end{itemize}

\subsection{Thought Node}
\label{sec:2.2}
\paragraph{Definition} 

A Thought Node corresponds to a distinct step or action within the CoT reasoning process. As outlined in Table~\ref{tab:prefill_prompt}, each node has a dedicated role and prefix prompt that guides the language model to generate specific content or revise previously generated reasoning. This structured design facilitates modular expansion and systematic backtracking within the MCTS framework. Notably, \textit{Thinking} is treated as a special node that does not require any prefix prompt; instead, it admits unconditioned continuation generation to foster open-ended exploration of partial solutions. By combining multiple node types into a coherent tree, we can more effectively elicit and refine multi-step reasoning from the model.

\begin{table}[t]
\centering
\scalebox{0.85}{
\begin{tabular}{l p{5.8cm}}
\toprule
\textbf{Thought Node} & \textbf{Prompt} \\
\midrule
Thinking & (continuation generation) \\
\hdashline\noalign{\vskip 0.5ex}
Sub-Task & Firstly, I need to break down this task. \\
\hdashline\noalign{\vskip 0.5ex}
Reflection & Let's check the result. Wait! something is wrong, let's think again. \\
\hdashline\noalign{\vskip 0.5ex}
Hypothesis & I propose the following hypothesis: \\
\hdashline\noalign{\vskip 0.5ex}
Double-Check & Now, I need to check whether all the requirements are met. \\
\hdashline\noalign{\vskip 0.5ex}
Reclarify & To ensure clarity, let me restate the question or issue at hand: \\
\hdashline\noalign{\vskip 0.5ex}
Answer & The answer is: \\
\bottomrule
\end{tabular}}
\caption{The pre-defined Thought Node. For a selected node, its corresponding prompt is continuously fed to LLMs for MCTS expansion.}
\label{tab:prefill_prompt}
\end{table}

\begin{figure*}[t]
\centering
\includegraphics[width=1.0\textwidth]{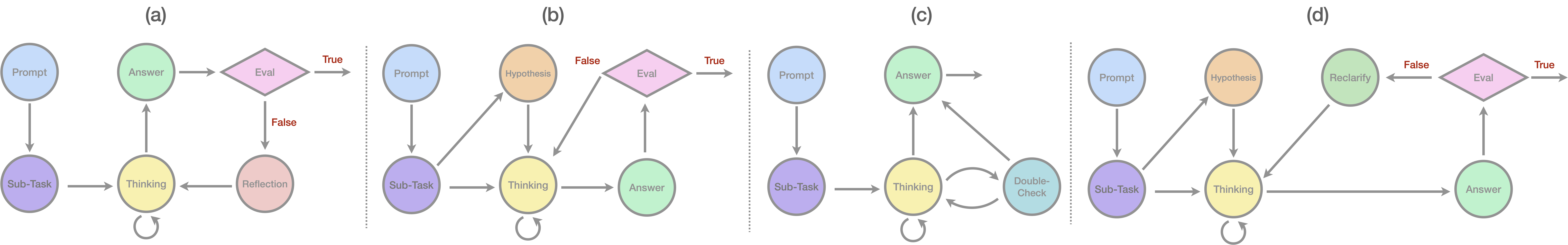}
\caption{Representative node transition patterns in our search tree. Each sub-figure (\textbf{a}--\textbf{d}) illustrates a distinct sequence of transitions (\emph{e.g.}, \textit{Sub-Task}, \textit{Thinking}, \textit{Reflection}, \textit{Double Check}, \textit{Hypothesis}) toward arriving at an \textit{Answer} node. These variations allow the search to adaptively expand or backtrack based on correctness checks, thereby generating rich and context-specific chain-of-thought data.}
\label{fig:MCTS_template}
\end{figure*}

\paragraph{Multi-Model Coordination and Reflection}
\label{sec:multi_model}

We adopt {multi-model coordination} to further diversify and correct the generated reasoning paths:
1) For nodes such as \textit{Thinking}, we use {Qwen2.5-72B-Instruct} to generate logical steps or partial solutions; 2) For \textit{Reflection} nodes, we switch to a different model, e.g., {Llama3.1-70B-Instruct}, to perform self-checks and corrections.

This separation enhances the reliability of reflection. When the same model that made a mistake also attempts to correct itself, it may fall back on the same erroneous distributional patterns. In our pipeline, if a \textit{Reflection} node detects an error, it can {backtrack} to a specific earlier node (also configurable in the MCTS design) and request a re-generation of the \textit{Thinking} steps. By alternating between models, we reduce the risk of repeated mistakes and improve the diversity of exploration.

\subsection{Reasoning Pattern}
\label{sec:2.3}
As illustrated in Figure~\ref{fig:MCTS_template}, we design a set of \emph{customizable search tree structures} to reflect the diverse ways in which humans reason about different tasks. Each tree is configured to capture a variety of reasoning modes. For instance, in Figure~\ref{fig:MCTS_template}(a), we demonstrate a sequence of nodes to solve a question: we first break down the task via a \textit{Sub-Task} node, then perform a general \textit{Thinking} step, and finally provide an \textit{Answer}. We evaluate correctness through rule-based checks: if correct, we output the result; otherwise, we prompt the model to reflect on potential mistakes (entering the \textit{Reflection} node). Here, the model revisits or re-checks its chain of thought, then either formulates new reasoning or proposes a revised answer. This feedback loop repeats until the solution is correct or a preset search limit is reached. 
Some tasks, however, also require formulating assumptions or provisional conclusions, so in Figure~\ref{fig:MCTS_template}(b), we incorporate a \textit{Hypothesis} node immediately after the \textit{Sub-Task} node for tasks that benefit from explicitly positing assumptions or preliminary formulas early on. For example, ``Find the sum of all ordered pairs \((x, y)\) of positive integers such that \(x + y = 5\).''  After breaking down the task (\textit{Sub-Task}), the model proposes a hypothesis that \(x\) can range from 1 to 4 (with \(y=5-x\)) and enumerates each pair, obtaining a total sum of 20. If the answer is verified correct, the model outputs 20.

In practice, we randomly sample from these different reasoning-flow templates (e.g., Figure~\ref{fig:MCTS_template}(a-d)) to ensure we capture diverse, human-like modes of thought. By introducing node transition patterns, our framework produces richer and more flexible CoT data and fosters more robust reasoning in downstream tasks.

\begin{figure*}[t]
\centering
\includegraphics[width=\textwidth]{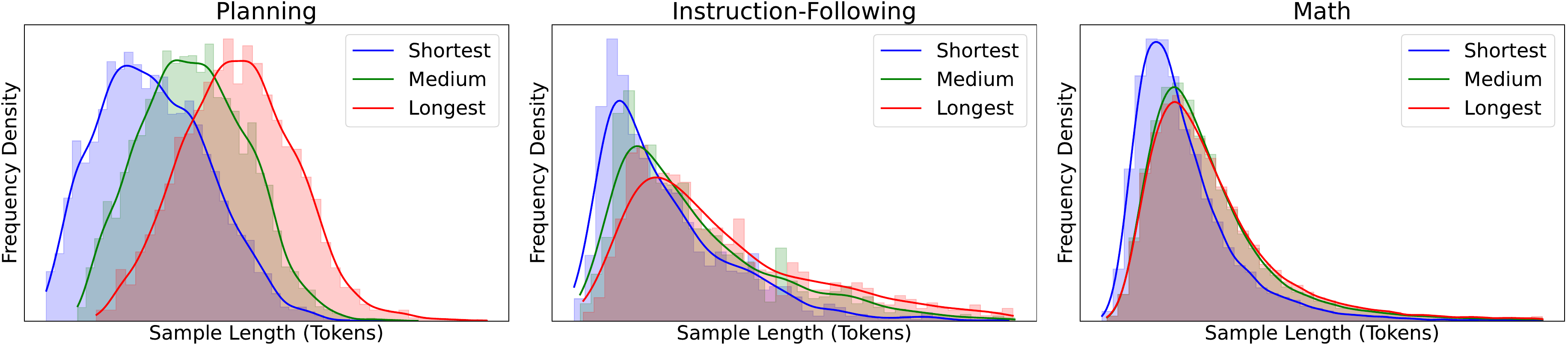}
\caption{Distribution of data lengths by token count: A comparative analysis of sampling strategies and their correspondence to short, medium, and long data lengths illustrated with histograms and KDE curves}
\label{fig:token_length_distribution}
\end{figure*}

\subsection{CoT Data Extraction for Post-Training}
\label{sec:sft_dpo_data}

Once MCTS completes its exploration, we have a large set of candidate paths. At this point, we must extract final CoT data for SFT or DPO.
\begin{itemize}[leftmargin=*,topsep=0.1em,itemsep=0.1em,parsep=0.1em]
\item{\bf CoT Data for SFT.} We select successful paths that lead to the correct final answer. Depending on the data volume requirements, one can: 1) Pick the highest-reward path according to MCTS; 2) Pick the long or short path that yields the correct answer, if specific chain lengths are desired.

\item{\bf CoT Data for DPO.} Constructing DPO data requires both positive and negative examples for each prompt: 1) The positive example is the CoT path that correctly solves the problem, like the SFT data; 2) The negative example is a flawed path (an incorrect final answer) that shares a minimal prefix with the positive path to mitigate excessive overlapping tokens, which can degrade DPO performance.
\end{itemize}
We find that many existing QA prompts (especially those frequently seen during Qwen or Llama training) are too easy for the models, producing few or no negative paths. As a result, fewer DPO pairs are generated. One can overcome this limitation by using more challenging questions or those that the models have not encountered extensively.

\begin{table}[t]
\centering
\scalebox{0.85}{
\begin{tabular}{l ccc}
\toprule
Datasets  & \bf Long & \bf Middle & \bf Short \\
\midrule
GSM8K & \underline{5.38\%} & \bf 5.08\% & \bf 5.08\% \\
MATH & 28.40\% & \underline{20.60\%} & \bf 16.20\% \\
AIME & \underline{51.66\%} & \bf 50.00\% & 60.00\% \\
\hdashline\noalign{\vskip 0.5ex}
Plan. & \underline{6.40\%} & \underline{6.40\%} & \bf 6.20\% \\
\hdashline\noalign{\vskip 0.5ex}
IF (Zh) & 32.30\% & \underline{4.23\%} & \bf 1.9\% \\
IF (En) & 22.36\% & \bf 3.80\% & \underline{4.00\%} \\
IF (Ot.) & 18.69\% & \underline{1.70\%} & \bf 1.59\% \\
\bottomrule
\end{tabular}} 
\caption{Effects of thoughts length(long, medium, and short CoT paths) on model performance across different datasets.}
\label{tab:data_balance}
\end{table}

\paragraph{Quantitative Analysis: Formalistic Long-time Thinking}
We explore the impact of the CoT length distribution on model performance. We experiment with different sampling strategies for DPO datasets, where responses to the same question are ranked and categorized based on their length (Their distributions are shown in Figure~\ref{fig:token_length_distribution}). As shown in Table~\ref{tab:data_balance}, the selection of shorter CoT paths leads to a noticeable reduction in ineffective outputs. In addition, we note that shorter reasoning paths tend to mitigate the issue of "formalistic long-time thinking", thus improving the quality of the reasoning output.

\section{CoT-Aware Post-Training}
\label{CoT-Aware Post-Training}
\label{sec:3}

Section 2.4 identifies that DPO training is prone to causing the formalistic long-time thinking. 
In this section, we propose three methods to address this problem: Thoughts Length Balance (Section \ref{sec:3-1}), Fine-grained DPO (Section \ref{sec:3-2}), Joint Post-training Objective (Section \ref{sec:3-3}).

\subsection{Thoughts Length Balance}
\label{sec:3-1}


Section 2.4 illustrates that the length of Chain-of-Thought (CoT) reasoning significantly influences the reasoning performance of distilled smaller models during the Direct Preference Optimization (DPO) phase. In contrast, our preliminary experiments did not indicate such an effect during SFT. Therefore, we propose using the longest CoT data in the SFT phase and the shortest CoT data in the DPO phase.

Specifically, we extract all valid reasoning paths that lead to correct answers from the CoT trees, as multiple correct paths often exist. From these, we select paths categorized by their relative length(short, medium, and long) based on token count, correct paths serve as positive examples, whereas negative examples are generated by identifying incorrect paths that share the shortest common prefix with their corresponding positive paths.

We recognize that "short" and "long" are inherently relative and that CoT lengths vary significantly with problem complexity. For instance, a simple arithmetic problem like "1+1=?" naturally involves a shorter CoT compared to more complex integrals such as "$\int_0^1 \frac{\ln (x+1)}{x^2+1},dx=?$". To address these variations systematically, we employ Monte Carlo Tree Search (MCTS) to sample multiple reasoning paths per query and select representative paths by their relative lengths, rather than imposing rigid token-count thresholds. 

\begin{figure}[t]
\centering
\includegraphics[width=0.5\textwidth]{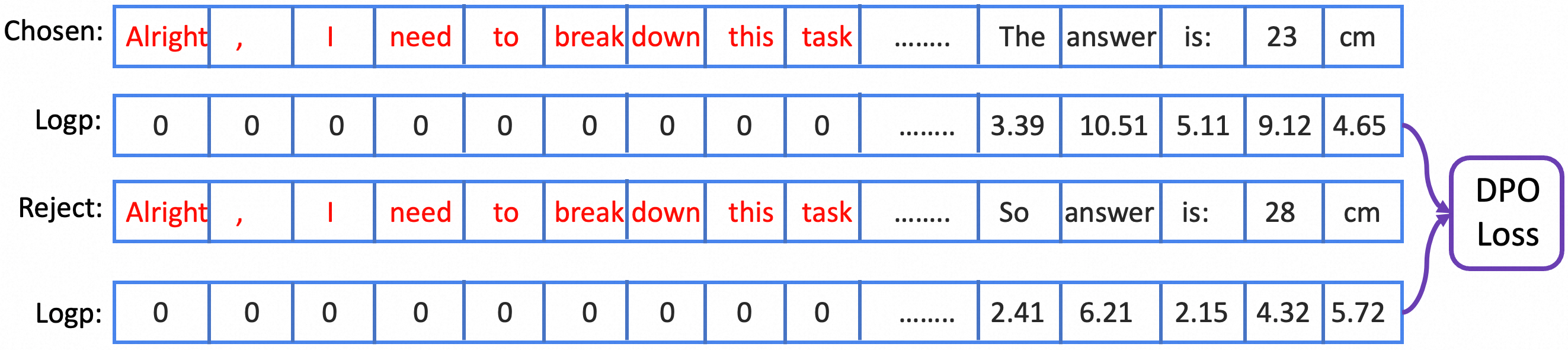}
\caption{The illustration of masking-based DPO, by setting the log probabilities of the common prefix tokens in preference pairs to zero.}
\label{fig:masking}
\end{figure}

\subsection{Fine-grained DPO}
\label{sec:3-2}


Recent studies highlight that DPO is sensitive to the length of responses, which can lead to biased reward assessments \cite{lu2024eliminatingbiasedlengthreliance,liu2024lengthdesensitizationdirectpreference}. 
Longer chosen responses increase the model's tendency to generate longer outputs, while longer rejected responses push the model to move away from such outputs, potentially without reducing their length. These issues are particularly pronounced in long CoT reasoning tasks, where length disparities may undermine DPO's effectiveness in fine-tuning reasoning models.


\begin{table*}[t]
  \centering
  \scalebox{0.85}{
    \begin{tabular}{l ccc c ccc}
      \toprule
      \multirow{2}{*}{\bf Model} & \multicolumn{3}{c}{\bf Math} &
      \multicolumn{1}{c}{\bf Planning} &
      \multicolumn{3}{c}{\textbf{Instruction-Following}} \\
      \cmidrule(lr){2-4} \cmidrule(lr){5-5} \cmidrule(lr){6-8}
        & \textbf{GSM8K} & \textbf{MATH} & \textbf{AIME} &
        \textbf{Blocksworld} &
        \textbf{Zh} & \textbf{En} & \textbf{Other} \\
      \midrule
      \textbf{Llama-3.1-8B-Instruct}      & 85.5 & 47.0 & 11.7 & 10.0 & 61.5 & 76.2 & 67.1 \\
      \textbf{~~+ Sky-T1}   & 84.8 & 44.0 & 6.7  & 2.0 & 25.4 & 31.6 & 29.7 \\
      \textbf{~~+ Our Data} & \bf 87.4 & \bf 51.4 & \bf 15.0 & \bf 12.4 & \bf 69.2 & \bf 76.6 & \bf 79.1 \\
      \midrule
      \textbf{Llama-3.2-1B-Instruct}      & 43.4 & 33.0 & 0.0 & 0.2 & 31.9 & 47.8 & 33.1 \\
      \textbf{~~+ Sky-T1}   & 37.2 & 19.0 & 0.0 & 0.0 &  8.1 & 10.0 &  7.5 \\
      \textbf{~~+ Our Data} & \bf 50.0 & \bf 38.6 & \bf 0.0 & \bf 5.6 & \bf 41.1 & \bf 50.6 & \bf 47.1 \\
      \midrule
      \textbf{Qwen2.5-7B-Instruct}      & 90.4 & 62.0 & 15.0 & 10.6 & 69.6 & 72.8 & 74.4 \\
      \textbf{~~+ Sky-T1}   & 89.6 & 61.6 & 9.4 & 0.4 &  26.2 & 24.5 & 30.6 \\
      \textbf{~~+ Our Data} & \bf 90.7 & \bf 64.0 & \bf 15.0 & \bf 12.0 & \bf 73.1 & \bf 73.4 & \bf 78.8 \\
      \midrule
      \textbf{Qwen2.5-1.5B-Instruct}      & 67.5 & 38.4 & 0.0 & 1.0 & 49.6 & 43.8 & 41.0 \\
      \textbf{~~+ Sky-T1}   & 66.7 & 34.4 & 0.0 & 0.0 &  9.2 & 11.8 & 14.8 \\
      \textbf{~~+ Our Data} & \bf 74.6 & \bf 46.8 & \bf 0.0 & \bf 5.4 & \bf 51.5 & \bf 51.6 & \bf 54.8 \\
      
      \bottomrule
    \end{tabular}
  }
  \caption{Comparison of SFT results across various models on multiple different tasks, including math, planning and instruction-following Benchmarks.}
  \label{tab:SFT_results}
\end{table*}

\paragraph{Conservative DPO} 

Conservative DPO (cDPO) \cite{mitchell2023note} adapts the standard DPO framework to handle noisy preference labels, typically encountered when labels may be flipped with a small probability $\epsilon$. The key innovation of cDPO is to modify the target distribution to account for potential label noise, setting the preference probability to $p(y_{w} \succ y_{l}) = 1 - \epsilon$. This adjustment reduces the impact of noisy labels by softening the gradient updates, making the model less sensitive to incorrect preferences. Formally, the cDPO loss is defined as:
\[
\begin{aligned}
\mathcal{L}_{\mathrm{DPO}}^{\epsilon}(\theta, y_w, y_l) = &-(1 - \epsilon) \log \hat{p}_{\theta}(y_w \succ y_l) \\
&- \epsilon \log(1 - \hat{p}_{\theta}(y_w \succ y_l)),
\end{aligned}
\]
where $\hat{p}_{\theta}(y_w \succ y_l)$ is the predicted preference probability. The gradient of this loss combines weighted contributions from both the correct and incorrect paths, facilitating more stable training under label noise. By upweighting correct preferences and downweighting incorrect ones, cDPO prevents overfitting to noisy data, resulting in more reliable optimization and improved model robustness.

\paragraph{Masking-based DPO} 

To mitigate the adverse effects of shared prefixes inherent in tree-based data, we modify the DPO loss computation by masking out the shared prefix tokens. Specifically, prior to loss calculation, we identify the number of tokens constituting the common prefix between the positive and negative samples and adjust the loss mask by setting the corresponding entries for these shared tokens to zero-analogous to the treatment of padding tokens in standard loss formulations, as shown in Figure~\ref{fig:masking}. This ensures that the shared prefix tokens do not contribute to the gradient computation, allowing the model to focus on the differentiating segments of the outputs and better distinguish between valid and invalid reasoning paths. This strategy provides a fine-grained adjustment to the DPO objective, enhancing optimization in settings with substantial prefix overlap.


\subsection{Joint Post-training Objective}
\label{sec:3-3}




In model training, the typical approach follows a sequential training paradigm, first conducting SFT followed by RLHF or DPO. However, this sequential training process is suboptimal due to the inherent trade-off between SFT and RLHF/DPO, where the model tends to forget the content learned in the first stage as it progresses to the second. Even regularization methods like KL divergence cannot fully mitigate the forgetting caused by the distribution shift from the SFT dataset to the preference-based dataset, as highlighted by \cite{fernando2025mitigatingforgettingllmsupervised}. A similar phenomenon is observed in our work, where such forgetting contributes to the emergence of formalistic long-time thinking in distilled models. To address this, we introduce SFT loss during the DPO training phase to alleviate the performance degradation resulting from the switch in training methodologies. The final loss function is thus modified as:
$\mathcal{L} = \mathcal{L}_{\text{DPO}} + \alpha \, \mathcal{L}_{\text{SFT}}$,
where the hyperparameter \(\alpha\) enables a better trade-off between SFT and preference learning, helping to maintain consistency in the model's reasoning patterns throughout the training process. This adjustment ensures more robust and stable performance across stages.

%

\section{Experiments}
\label{sec:4}

\subsection{Experimental Setup}

\paragraph{Models}

We start with the baseline model, "Our LRM (SFT)," which is Llama-3.1-8B fine-tuned on our CoT data. Direct Preference Optimization (DPO) is applied next, followed by Data Length Balance. Conservative DPO (cDPO) is then added, and a Joint Loss function combining DPO and Supervised Fine-Tuning (SFT) loss is incorporated. Finally, masking-based DPO is applied. Each of these methods is sequentially added to the baseline, as shown in Table~\ref{tab:performance}.

\begin{table*}[t]
\centering
\scalebox{0.85}{
\begin{tabular}{lccccccc}
\toprule
\multirow{2}{*}{\bf Model} & \multicolumn{3}{c}{\bf Math}  & \multicolumn{1}{c}{\bf Planning} & \multicolumn{3}{c}{\textbf{Instruction-Following}} \\
\cmidrule(lr){2-4} \cmidrule(lr){5-5} \cmidrule(lr){6-8}
  & \textbf{GSM8K} & \textbf{MATH} & \textbf{AIME} & \textbf{Blocksworld} & \textbf{Zh} & \textbf{En} & \textbf{Other} \\
\midrule
\multicolumn{8}{c}{\em Baseline}\\
Our LRM (SFT) & \underline{87.4} & \bf 51.4 & \bf 15.0  & \underline{12.4} & 69.2 & 76.6 & \bf 79.1 \\
& ({\underline{\em 0.23\%}}) & ({\em\bf 5.40\%}) & ({\em\bf 30.00\%}) & ({\em\bf 1.80\%})& ({\em\bf 0.77\%}) & ({\em\bf 1.69\%}) &({\em\bf 1.08\%}) \\
\hdashline\noalign{\vskip 0.5ex}
\multirow{2}{*}{~~+ DPO} & 86.2 & 41.8 & 8.3 & 2.0 & 5.7 & 6.3 & 6.7  \\ 
& ({\em 6.37\%}) & ({\em 31.80\%}) & ({\em 55.00\%}) & ({\em 93.60\%})& ({\em 91.54\%}) & ({\em 90.93\%}) &({\em 92.22\%}) \\
\midrule
\multicolumn{8}{c}{\em Our Methods}\\
\multirow{2}{*}{~~+ Data Balance} & 86.8 & 28.0  & 6.6 & 6.8 & 43.4  & 44.7 & 42.4 \\
& ({\em 5.08\%}) & ({\em 46.40\%}) & ({\em 65.00\%}) & ({\em 44.60\%})& ({\em 30.77\%}) & ({\em 44.73\%}) &({\em 45.28\%}) \\
\hdashline\noalign{\vskip 0.5ex}
\multirow{2}{*}{~~~~+ cDPO} & \bf 87.5 & 48.6 & \bf 15.0  & 4.4 & 61.9 & 66.4 & 67.7 \\
& ({\em 3.71\%}) & ({\em 15.00\%}) & ({\em 45.00\%}) & ({\em 47.40\%})& ({\em 11.15\%}) & ({\em 15.61\%}) &({\em 15.40\%}) \\
\hdashline\noalign{\vskip 0.5ex}
\multirow{2}{*}{~~~~~~+ Joint Loss} & 86.8 & 48.6 & \underline{10.0}   & 8.6 & \bf 72.3 & \bf 78.9 & \underline{78.1} \\
& ({\em 0.38\%}) & ({\em 8.60\%}) & ({\underline{\em 31.67\%}}) & ({\underline{\em 9.00\%}})& ({\underline{\em 1.15\%}}) & ({\underline{\em 1.90\%}}) &({\em 2.22\%}) \\
\hdashline\noalign{\vskip 0.5ex}
\multirow{2}{*}{~~~~~~~~+ Masking} & 87.2 & \underline{51.0}  & 8.0 & \bf 12.6 & \underline{72.0} & \underline{77.2} & \bf 79.1 \\
& ({\em\bf 0.15\%}) & (\underline{{\em 5.80\%}}) & ({\em 38.33\%}) & ({\em 10.20\%})& ({\underline{\em 1.15\%}}) & ({\underline{\em 1.90\%}}) &({\underline{\em 1.36\%}}) \\
\bottomrule
\end{tabular}%

}
\caption{Performance comparison among different methods. The best performance is boldfaced, while the second best is underlined. The numbers in parentheses indicates the ratio of instances where no answer is obtained in the specified format.}
\label{tab:performance}
\end{table*}

\paragraph{Benchmark}

We evaluate our approach on five benchmarks, each capturing different reasoning challenges. AIME focuses on higher-level math with 60 questions from 2023 and 2024, while GSM8K~\cite{cobbe2021gsm8k} features elementary-to-intermediate arithmetic tasks. MATH500~\cite{lightman2023lets} presents a wide range of advanced mathematical problems, testing deeper analytical thinking. For sequential decision making, we adopt the classical Blocksworld~\cite{Valmeekam2022PlanBenchAE} planning domain from the International Planning Competitions (IPC). Lastly, Multi-IF~\cite{he2024multi} assesses multi-turn instruction following in eight languages, encompassing 4,501 multilingual, three-turn conversations.

\subsection{Experimental Results}

\paragraph{Constructed Data Validation} We apply our constructed CoT data to smaller-scale models from the Llama and Qwen families, comparing the results against the Sky-T1 dataset, which employs a QwQ-based distillation pipeline effective on larger models (32B). While Sky-T1 demonstrates competitive performance on large models, it faces challenges when scaled down to 8B models due to the inherent limitations in context processing and reasoning capabilities. In contrast, our CoT data, specifically designed to address these limitations, leads to substantial improvements in smaller models, particularly in tasks involving arithmetic reasoning such as GSM8K and MATH, as well as more complex open-ended tasks like AIME and Blocksworld, as shown in Table~\ref{tab:SFT_results}.
This validates the effectiveness of our constructed data in advancing the performance of small models across a wide range of reasoning tasks.

\paragraph{Main Results}

As shown in Figure~\ref{tab:performance}, we progressively adding various techniques described in Section~\ref{CoT-Aware Post-Training} to address the challenges identified during DPO. Initially, we observe that DPO causes a significant increase in output length, which results in a marked drop in model performance due to the high proportion of samples without answers. To mitigate this issue, we explore several strategies, including the data balance, applying cDPO to reduce the impact of noisy labels, SFT Loss for multi-objective training to prevent catastrophic forgetting, and masking shared prefixes during loss calculation to reduce overemphasis on redundant tokens. 
Our results show that these adjustments achieves a notable improvement in reasoning tasks, particularly in planning and instruction-following, while maintaining competitive performance on mathematical benchmarks. The performance improvements can be primarily attributed to the reduction of formalistic long-time thinking, which is a major source of inefficiency in reasoning. By addressing this issue, our model exhibits a stronger ability to generate meaningful and correct answers instead of producing excessive, irrelevant reasoning steps. This leads to a substantial enhancement in overall model effectiveness, with improvements of coherent reasoning and accurate outputs. The integration of these methods ensures that our model achieves robust and efficient performance across a wide range of reasoning tasks, Contributing to the application of DPO technology in LRMs.

\subsection{Effects of Joint DPO Loss}

In this study, we explore the effects of combining DPO loss with SFT loss within a joint post-training objective, as outlined in Section 3.4. Our goal is to stabilize the training process and mitigate issues like over-optimization observed in pure DPO. To this end, we experiment with varying values of the hyperparameter alpha, which controls the weight balance between the DPO and SFT losses. As shown in Table~\ref{tab:joint_loss}, our results indicate that alpha=1 provides the best trade-off, as smaller values still lead to some degree of catastrophic forgetting, while larger values reduce the effectiveness of preference alignment, thereby diminishing the efficiency of the valuable preference dataset. Consequently, we adopt the combined $\mathcal{L} = \mathcal{L}_{\text{DPO}} + \mathcal{L}_{\text{SFT}}$ as the configuration for subsequent experiments.

\begin{table}[t]
\centering
\scalebox{0.68}{
\begin{tabular}{l c cccc}
\toprule
Datasets & \bf CDPO & \bf +0.5 & \bf +1.0 & \bf +1.5& \bf +2.0 \\
\midrule
\multirow{2}{*}{ GSM8K } &  \bf 87.5&86.5&\underline{86.8}&85.5&85.6 \\
& ({\em 3.71\%}) & ({\em 0.53\%}) & ({\underline{\em 0.38\%}}) & ({\em\bf 0.08\%})& ({\em\bf 0.08\%})  \\
\multirow{2}{*}{ MATH }& \underline{ 48.6}&\bf 50.0&\underline{48.6}&48.4&48.0\\
& ({\em 15.00\%}) & ({\em 10.20\%}) & ({\underline{\em 8.60\%}}) & ({\em\bf 0.08\%})& ({\em 9.00\%})  \\
\multirow{2}{*}{ AIME }&  \bf 15.0&\underline{11.6}&10.0&6.6&\underline{11.6}\\
& ({\em 45.00\%}) & ({\underline{\em 35.00\%}}) & ({\em\bf 31.67\%}) & ({\em 36.67\%})& ({\em\bf 31.67\%})  \\
\hdashline\noalign{\vskip 0.5ex}
\multirow{2}{*}{ Plan.} &  4.4&7.8&\bf 8.6&7.6&\underline{8.4}\\
& ({\em 47.40\%}) & ({\em 12.80\%}) & ({\em 9.00\%}) & ({\em\bf 6.40\%})& ({\underline{\em 7.40\%}})  \\
\hdashline\noalign{\vskip 0.5ex}
\multirow{2}{*}{ IF (Zh) }&  61.9&68.8&\bf 72.3&68.4&\underline{70.7}\\
& ({\em 11.15\%}) & ({\em 2.31\%}) & ({\underline{\em 1.15\%}}) & ({\em 3.46\%})& ({\em\bf 0.77\%})  \\
\multirow{2}{*}{ IF (En) }&  66.4&76.1&\bf 78.9&77.2&\underline{ 78.2}\\
& ({\em 15.61\%}) & ({\em 4.22\%}) & ({\em\bf 1.90\%}) & ({\em 2.74\%})& ({\underline{\em 2.32\%}})  \\
\multirow{2}{*}{ IF (Ot.) }&  67.7&78.0&78.1&\bf 79.2&\underline{78.9}\\
& ({\em 15.40\%}) & ({\em 1.99\%}) & ({\em 2.22\%}) & ({\em\bf 1.82\%})& ({\underline{\em 1.93\%}})  \\
\bottomrule
\end{tabular}}
\caption{Effects of joint loss (combining DPO loss and SFT loss with a weight factor $\alpha$ in $\mathcal{L} = \mathcal{L}_{\text{DPO}} + \alpha  \mathcal{L}_{\text{SFT}}$) on model performance across different datasets with varying hyperparameter settings.}
\label{tab:joint_loss}
\end{table}

\subsection{MCTS Inference Exploration}
\label{sec:appendix}

As an additional experiment, we explored the impact of applying MCTS at the inference stage. As shown in Table~\ref{tab:model_comparison_mstc}, we use Test@N to denote the percentage of problems solved correctly at least once when allowing the model to make N separate guesses for each problem \cite{cobbe2021trainingverifierssolvemath}. We evaluated solve rates at Test@1, Test@8, and Test@32 on the MATH dataset. Specifically, at Test@8 and Test@32, the MCTS-based approach outperforms the model without MCTS inference, demonstrating its ability to expand the solution space and leverage test-time scaling effectively.

\section{Related Work}

\paragraph{Reasoning Models}

Recent advancements in reasoning models, like OpenAI o1 \cite{openai2024reason}, DeepSeek-R1 \cite{deepseekai2025deepseekr1incentivizingreasoningcapability}, and Qwen QwQ \cite{qwq-32b-preview}, make significant strides in complex reasoning tasks through CoT generation and increased test-time compute. These models scale reasoning depth by expanding their thinking processes, generating step-by-step solutions to problems, which significantly boosts performance in domains such as mathematics and coding \cite{zhong2024evaluationopenaio1opportunities, deepseekai2025deepseekr1incentivizingreasoningcapability}. However, the challenge remains that these advancements largely depend on large model sizes, and smaller models struggle to replicate this reasoning behavior \cite{fu2023specializingsmallerlanguagemodels}.

\paragraph{Knowledge Distillation}

Researchers focus on transferring the reasoning capabilities of LRMs into smaller models, through distillation techniques. Distillation methods, such as fine-tuning smaller models on CoT data generated by larger models \cite{huang2024o1replicationjourney}, show that small models can benefit from reasoning data generated by large models. However, recent findings reveal that small models often fail to capture intricate reasoning patterns due to their limited capacity, leading to suboptimal performance when directly distilled from large models \cite{li2025smallmodelsstrugglelearn, wang2025thoughtsplaceunderthinkingo1like}. 

\paragraph{Monte Carlo Tree Search}

MCTS has been proposed as a solution to improve reasoning by exploring multiple reasoning paths during inference \cite{zhao2024marcoo1openreasoningmodels, tian2024selfimprovementllmsimaginationsearching}. Moreover, MCTS provides a powerful mechanism for generating high-quality, diverse reasoning data that can subsequently be harnessed to fine-tune reasoning models. For instance, \cite{tian2024selfimprovementllmsimaginationsearching} employs MCTS to synthesize candidate reasoning paths that capture varied solution strategies. Similarly, the RStar introduced in \cite{qi2024mutualreasoningmakessmaller} utilizes MCTS to construct structured data, ensuring that generated reasoning chains are both comprehensive and coherent. In parallel, Math-Shepherd \cite{wang2024mathshepherdverifyreinforcellms} and OmegaPRM \cite{luo2024improvemathematicalreasoninglanguage} employ MCTS to collect high-quality reasoning data, which is subsequently used to train PRM. Extending these ideas, \cite{zhang2024restmctsllmselftrainingprocess} combines the strengths of MCTS and PRM to guide policy updates.

\begin{table}[t]
\centering
\scalebox{0.85}{
\begin{tabular}{l ccc}
\toprule
Model & \bf Test@1 & \bf Test@8 & \bf Test@32 \\
\midrule
Llama-3.1-8B-Instruct & 47.0 & 67.6 & \bf 75.8 \\
Our Best Model & 51.0 & 70.2 & \bf 79.2 \\
\midrule
~~+ MCTS Decode & 51.0 & 70.8 & \bf 82.8 \\
\bottomrule
\end{tabular}}
\caption{Performance on MATH Dataset: Test@1, Test@8, and Test@32 Results. Test@N denotes the percentage of problems solved correctly at least once when the model is allowed to make N separate guesses for each problem.}
\label{tab:model_comparison_mstc}
\end{table}

\section{Conclusion}

We explore strategies to transfer long CoT reasoning to smaller models, addressing learning difficulty and bias inheritance in distillation. We propose a MCTS framework that independently generates flexible, tree-based reasoning data, reducing reliance on large teacher models. Enhanced by CoT-aware post-training, our approach effectively mitigates overly formalistic long-time thinking. Experiments across diverse tasks (math, planning and
instruction-following) demonstrate robust performance improvements. These findings highlight the importance of well-designed data and post-training strategies in improving the efficiency and reliability of smaller-scale reasoning models.
In the future, we will continue to explore how reasoning techniques can improve a broader range of non-mathematical tasks, including machine translation \cite{liu2025new}, multilingual and multimodal reasoning \cite{zeng2025marco,li2025perception}.

\bibliography{main}





\end{document}